\documentclass[final]{cvpr}

\usepackage{times}
\usepackage{epsfig}
\usepackage{graphicx}
\usepackage{amsmath}
\usepackage{amssymb}
\usepackage{amsfonts}
\usepackage{tabularx}
\usepackage{algpseudocode}
\usepackage{algorithm}

\usepackage[pagebackref=true,breaklinks=true,colorlinks,bookmarks=false]{hyperref}

\def\cvprPaperID{****} 
\def\confYear{CVPR 2021}

\begin{document}

\title{Improving Ranking Correlation of Supernet with Candidates Enhancement and Progressive Training}

\author{Ziwei Yang\textsuperscript{1}, Ruyi Zhang\textsuperscript{1}, Zhi Yang\textsuperscript{1}, Xubo Yang\textsuperscript{1}, Lei Wang\textsuperscript{3} and Zheyang Li\textsuperscript{1,2}\\
\textsuperscript{1}Hikvision Research Institute, \textsuperscript{2}Zhejiang University \\
\textsuperscript{3}University of Science and Technology of China\\
{\tt\small \{yangziwei5,zhangruyi5,yangzhi13,yangxubo,lizheyang\}@hikvision.com} \\
{\tt\small wangl26@mail.ustc.edu.cn} 
}

\maketitle

\begin{abstract}

One-shot 
neural architecture 
search (NAS) applies 
weight-sharing supernet 
to reduce the 
unaffordable computation overhead of automated architecture designing.
However, the weight-sharing technique worsens the ranking consistency of performance due to the interferences between different candidate networks.
To address this issue, we propose a candidates enhancement method and progressive training pipeline to improve the ranking correlation of supernet.
Specifically, we carefully redesign the sub-networks in the supernet and map the original supernet to a new one of high capacity. 
In addition, we gradually add narrow branches of supernet to reduce the degree of weight sharing which effectively alleviates the mutual interference between sub-networks.
Finally, our method ranks the 1st place in the Supernet Track of CVPR2021 1st Lightweight NAS Challenge.
Our code is released on \url{https://github.com/Tend93/CVPR2021\_NAS\_competition\_Track1\_1st\_solution}.

\end{abstract}

\section{Introduction}

Neural architecture search aims to automatically design neural architectures. 
Although the architectures found by NAS methods \cite{cai2019ofa,chu2019scarlet,Zhou_2020_ecosnas} outperform human designed ones in many computer vision tasks, 
the previous methods \cite{mnasfpn,tan2019mnasnet} require evaluting an enormous amount of candidate architectures with stand-alone training.
The unaffordable computation overhead of the evaluation leads to a weight-sharing strategy of NAS\cite{chu2019scarlet,guo2020spos}.

As one of the widely-used weight-sharing methods, 
one-shot NAS utilizes a supernet subsuming all candidate architectures within the search space to evaluate accuracies on the validation dataset.
Instead of stand-alone training, all architectures directly inherit their weights from the supernet which is only trained once.
The computation cost of evaluation for architectures is effectively reduced.

To train supernet, NAS methods \cite{chu2019fairnas, guo2020spos} sample one or a few sub-networks from supernet and train them in each update step.
Due to the weight-sharing fashion, sub-networks interfere with each other and exhibit inferior accuracies compared with strand-alone training.
The accuracy ranking of sub-networks evaluated by the weights inheriting is inconsistence with the accuracy ranking of sub-networks 
when they are trained from scratch independently\cite{cai2019ofa,chu2019scarlet}.

For this issue, recent NAS methods attempt to improve the ranking correlation of supernet from two perspectives: 
optimizing the training process of supernet and enhancing the capacity of supernet by sub-network redesigning.
FairNAS \cite{chu2019fairnas} employs a fair sampling strategy of sub-networks to increase the training accuracy.
OFA \cite{cai2019ofa} narrows the accuracy gap of sub-networks between supernet evaluating and stand-alone training by knowledge distilling.
Although these methods can improve the ranking correlation of supernet in regular search spaces,
it is difficult for training an irregular search space which contains searchable options with extreme conflicts, 
such as channel options 4 and 64.
Furthermore, SCARLET-NAS\cite{chu2019scarlet} adds extra weights to improve the capacity of supernet for better convergence and less interferences between sub-networks.
Laube et al.\cite{laube2021inter} trains the sub-networks independently by adding bias weights and splitting path weights.
These methods alleviate the disturbances of sub-networks, but the additional weights need carefully designs and hyper-parameters tuning.

\begin{figure*}[htbp]
   \begin{center}
   \includegraphics[height=0.4\textwidth]{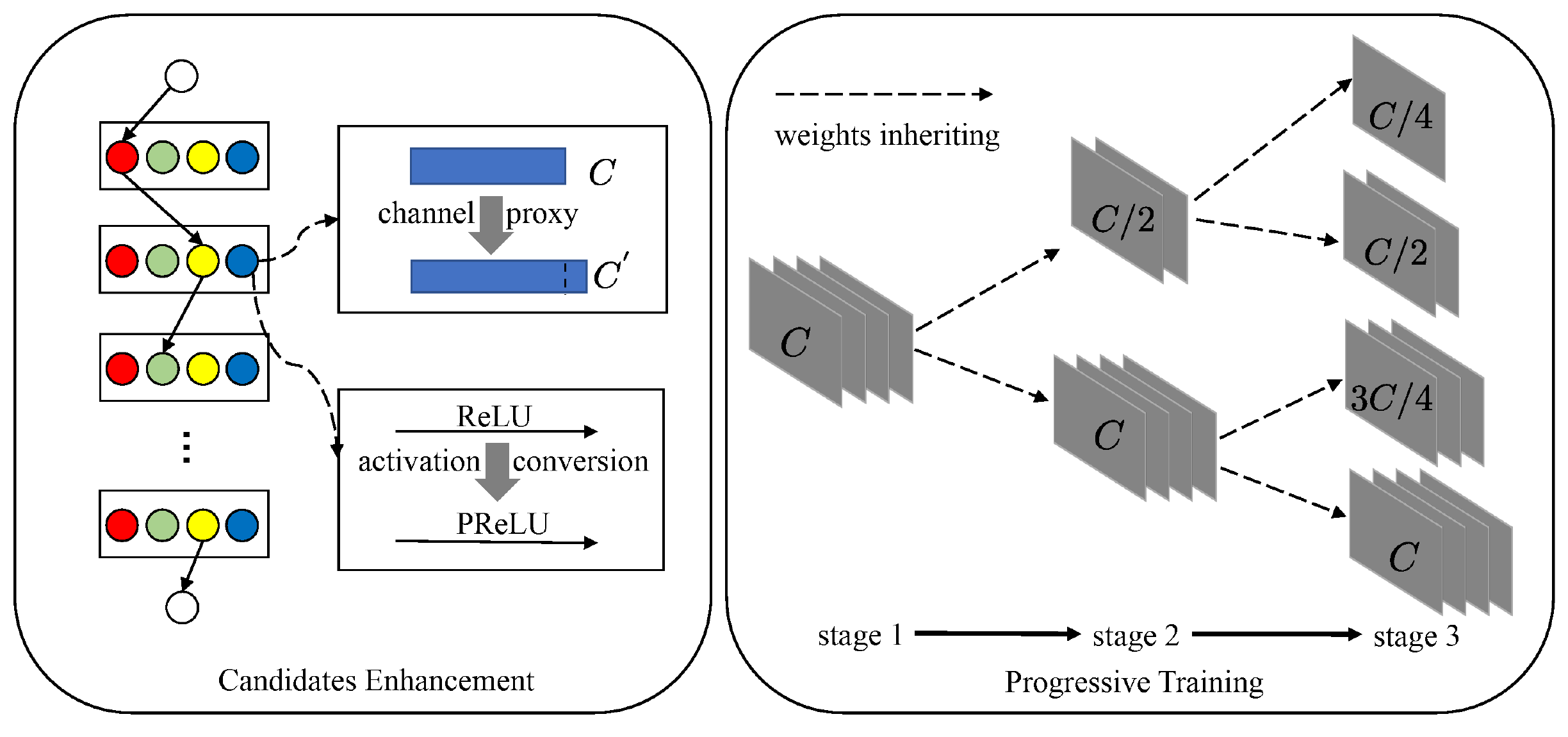}
   \end{center}
      \caption{The framework of our method. The left part displays the candidates enhancement of supernet.
      The ``blue'' option is enhanced by channel proxy and activation conversion. The corresponding sub-networks are strengthened. 
      The right part takes weights of one layer as example to show the progressive training of supernet.
      The weights are duplicated step by step for further finetuning}
   \label{fig:framework}
\end{figure*}

In this paper, we apply candidates enhancing and progressive training pipeline to improve the ranking consistency 
of candidate architectures evaluated via supernet and stand-alone training. 
We find some sub-networks in the supernet exhibit inferior accuracies in the joint optimization of all sub-networks.
It contributes to the degradation of ranking correlation between supernet evaluating and stand-alone training.
To tackle this problem, we convert the sub-networks to the ones of higher capacity, 
which only adds few weights but easy to train without extra hyper-parameters tuning. 
Besides, we utilize a progressive training method to alleviate the interference of sub-networks. 
For easily launching the training process, we firstly train the supernet by a variant of \cite{cai2019ofa}. 
Then we gradually append narrow branches of each searchable layer and finetune the weights.
The performance of our approach is certified in the \emph{Supernet Track of CVPR2021 1st Lightweight NAS Challenge} and ranks the 1st place.
\section{Method}

\subsection{Supernet Training with Distilling}\label{section:distilltrain}
We firstly introduce the algorithm of supernet training with knowledge distilling in our method.
The search space is denoted as $\mathcal{S}$ with weights $\theta$. 
It contains sub-networks $\left \{ s_{i} \right \} , i\in \left [1, N \right ]$.
$N$ is the number of all sub-networks in the supernet. $s_{N}$ represents the largest sub-network with maximum channels.

Supernet training with knowledge distilling can be formalized as 
\begin{center}
\begin{equation} \label{f1}
   \theta^{*} = \mathop{\arg\min}\limits_{\theta} \mathbb{E}_{s_{i}\sim \Gamma(\mathcal{S}) }  \left [\mathcal{L}\left (s_{i}, s_{N}, \theta \right ) \right ]
\end{equation}
\end{center}
where $\theta^{*}$ denotes the optimized weights of supernet.
$\mathcal{L}$ is the optimization function.
$\Gamma(\mathcal{S})$ is a sampling distribution of $s_{i} \in \mathcal{S}$.

We utilize cross entropy (CE) loss as a common optimization function for image classification task.
Then, Kullback–Leibler (KL) divergence loss is additionally applied for distilling sub-networks with the largest one.
The complete optimization function is defined as follows:
\begin{center}
\begin{equation}\label{f2}
   \begin{aligned}
   \mathcal{L} = \left(1-\alpha \right) &\mathcal{L}_{CE} \left( \mathcal{N} \left(s_{i}, \theta \right) \right)\\
   &+ \alpha \mathcal{L}_{KL}\left ( \mathcal{N}\left (s_{i}, \theta \right ), \mathcal{N}\left (s_{N}, \theta \right ) \right ),
   \end{aligned}
\end{equation}
\end{center}
where $\mathcal{L}_{CE}$ and $\mathcal{L}_{KL}$ denote the cross entropy loss and Kullback–Leibler divergence loss respectively.
$\mathcal{N} \left(s_{i}, \theta \right)$ represents the sub-network with architecture $s_{i}$ and inherited weights from $\theta$. 
$\alpha$ is a coefficient which trades off classification loss and distillation loss.
The training algorithm is illustrated in Algorithm~\ref{alg:algorithm1}.

\textbf{Search Space.} The search space is builded based on the ResNet20 backbone. 
The numbers of channels in each layer are searchable. 
The candidate channel options of layers are as follows:
\begin{itemize}
   \item layers 1 to 7: [ 4, 8, 12, 16].
   \item layers 8 to 13: [ 4, 8, 12, 16, 20, 24, 28, 32].
   \item layers 14 to 19: [ 4, 8, 12, 16, 20, 24, 28, 32, 36, 40, 44, 48, 52, 56,60, 64].
\end{itemize}
There are around $7.21*10^{16}$ sub-networks in total.

\subsection{Candidate Enhancing}
In order to alleviate the interference between sub-networks in the supernet,
we redesign the sub-networks for capacity enhancing.
We define a mapping function $\mathcal{R}$ to execute the transformation $s_{i} \rightarrow \mathcal{R}(s_{i})$.

In this paper, we redesign the sub-networks with two types of modification: activation conversion and options enhancement. 
The left part of Figure~\ref{fig:framework} shows details of the modification.
Firstly, we argue that the activation function ``ReLU'' is harmful for the small channels, 
which will filter too much information while the channels are only 4 or 8.
We change the activation function to ``PReLU'' which is more smooth.
In addition, the sub-networks in our search space within numerous channel options of 4 or 8 exhibit much lower accuracy in the supernet compared with stand-alone training.
This inspires us to enhance the capacity of channel options 4 and 8. 
We employ channel options 5 and 9 as proxies of options 4 and 8 respectively, 
which contribute to similar accuracy when they are trained stand alone. 
With the modifications, the accuracies of sub-networks are effectively increased and it makes a better accuracy ranking of sub-networks in the supernet.

\subsection{Progressive Training}
In the first stage of supernet training, 
we only maintain one copy of weights for different options in each layer as \cite{cai2019ofa} does and train the supernet by Algorithm~\ref{alg:algorithm1}.
Although this benefits the training of small sub-networks in the beginning, 
the mutual interference of sub-networks becomes more intense since the complete weight sharing.

Therefore, we insert extra counterparts of weights for each layer and allocate them to different searchable options,
which leads the original supernet to a multi-branch one.
The transformation and training of supernet are implemented step by step for elaborately supernet finetuning. 
To be specific, we duplicate the weights of layers in the supernet based on the pretrained weights of previous training stage.
The options of layers are divided into two parts and inherit different counterparts of the weights, 
which is exhibited in the right part of Figure~\ref{fig:framework}.
Then, we add more copy of weights for each layer and gradually reduce the number of options that share the same counterpart of weights.
With progressive training, sub-networks in the supernet can be elaborately finetuned, which leads to further improved accuracy with pretrained weights.

\begin{algorithm}[t]
   \caption{Supernet Training}
   \label{alg:algorithm1}
   \hspace*{0.02in}{\bf Input:}
   supernet $\mathcal{S}$ with weights $\theta$, trainging set $D_{train}$, 
   sampling number of sub-networks $K$, training iterations $T$, loss function $\mathcal{L}$.\\
   \hspace*{0.02in}{\bf Output:}
   optimized supernet weights $\theta^{*}$.
   \begin{algorithmic}[1]
      \State random initialize $\theta$,
      \State train largest sub-network of supernet with $\mathcal{L}_{CE}$.
      \For{$t \gets 1, T$}
         \State set gradients of all weights to zeros;
         \State forward largest sub-network $s_{N}$;
         \State calculate gradients based on $\mathcal{L}_{CE}$ .
         \For{$i \gets 1, K$}
            \State Sample sub-network $s_{i}$ by uniform sampling;
            \State forward sub-network $s_{i}$;
            \State distill the outputs of $s_{i}$ with the outputs of $s_{N}$;
            \State calculate gradients based on $\mathcal{L}$.
         \EndFor
         \State update $\theta$ by accumulated gradients.
      \EndFor
      \State return $\theta^{*}$.
   \end{algorithmic}
\end{algorithm}


\section{Experiments}
We evaluate our method in the \emph{Supernet Track of CVPR2021 1st Lightweight NAS Challenge}.
We firstly train the largest sub-network for 300 epoches with batch size $128$ and apply a stochastic gradient descent optimizer with a momentum of $0.9$.
The learning rate is decreased from an initial value of $0.1$ to $0$ with a cosine learning rate decay strategy. 
The weights are regularized with weight decay of $5e-4$. 
The data augmentation includes transformation of brightness and contrast, rotation of 15 degrees and random flipping.
We implement the progressive training with the same hyper-parameters of training as largest sub-network except a different learning rate.

\textbf{Metrics.} We follow the setting of CVPR2021 challenge and evaluate the ranking performance of supernet with the absolute value of the Pearson correlation coefficient. 
We test 50,000 sub-networks provided by the challenge with inherited weights of the well-trained supernet. 
The Pearson correlation coefficient is calculated with a part of the sub-networks. 

\subsection{Results of Candidates Enhancement}
We evaluate the effectiveness of candiates enhancement with one-stage supernet training illustrated in Algorithm~\ref{alg:algorithm1}.
The number of sub-networks sampling in each weight update step is set to $8$ and the initial learning rate of supernet training is $0.01$.

The experimental results are displayed in Table~\ref{table:enhancement}, 
where ``base'' represents the original supernet, ``$OE$'' means the supernet with options enhancement, ``$PRL\_OE$'' denotes the supernet with both options enhancement and activation conversion to ``PReLU''.
From the table, we find that the ranking correlation of $ResNet20\_SPN\_OE$ outperforms that of base supernet $ResNet20\_SPN$ by 0.006.
The score is further increased to $0.97321$ after combining options enhancement and activation conversion, 
indicating the availability of candidates enhancement in our method.

\begin{table}
   \begin{center}
   \scalebox{0.86}{
   \begin{tabular}{|c|c|c|}
   \hline
   Model & Supernet & Pearson Coeff. \\
   \hline\hline
   $ResNet20\_SPN$ & base & $0.96341$ \\
   $ResNet20\_SPN\_OE$ & OE & $0.96944$ \\
   $ResNet20\_SPN\_PRL\_OE$ & PReLU+OE & $0.97321$ \\
   \hline
   \end{tabular}}
   \end{center}
   \caption{The ranking correlation of sub-networks in the supernet. 
   ``OE'' represents the supernet with options enhancement.}
   \label{table:enhancement}
\end{table}

\subsection{Results of Progressive Training}
The progressive training is consisting of three stages. Firstly, we train the supernet with complete weight sharing, where all searchable options in one layer inherit the same weights.
Afterwards, the weights are duplicated and searchable options for each layer are divided into two sets, 
such as splitting [4,8,12,16] to [4,8] and [12,16].
Finally, the number of weights in the supernet is further increased. Each option will have an exclusive counterpart of weights.
The structures of supernet and training process are displayed in the right part of Figure~\ref{fig:framework}.
The supernet training of second stage is implemented based on the pretrained weights of first stage with an initial learning rate of $0.001$.
Third stage utilizes the same learning rate as second stage with latest pretrained weights.

Table~\ref{table:progressive} exhibits the experimental results of progressive training. 
The table demonstrates that second stage training makes a great improvement compared with the result of stage $1$ by 0.0039 in ``base'' supernet and $0.0033$ in ``PReLU+OE'' supernet. 
However, the training of third stage only works in supernet ``PReLU+OE'' because of the lower capacity of ``base'' supernet.

\begin{table}[htbp]
   \begin{center}
   \scalebox{0.7}{
   \begin{tabular}{|c|c|c|c|}
   \hline
   Model & Supernet & Training stage &Pearson Coeff. \\
   \hline\hline
   $ResNet20\_SPN$ & base & $1$ & $0.96341$ \\
   $ResNet20\_SPN$ & base & $2$ & $0.96732$ \\
   $ResNet20\_SPN$ & base & $3$ & $0.96686$ \\
   $ResNet20\_SPN\_PRL\_OE$ & PReLU+OE & $1$ & $0.97321$ \\
   $ResNet20\_SPN\_PRL\_OE$ & PReLU+OE & $2$ & $0.97648$ \\
   $ResNet20\_SPN\_PRL\_OE$ & PReLU+OE & $3$ & $0.97696$ \\
   \hline
   \end{tabular}}
   \end{center}
   \caption{The ranking correlation of supernet in different progressive-training stages.}
   \label{table:progressive}
\end{table}

\subsection{Visualization of ranking correlation}
\begin{figure}[htbp]
   \begin{center}
   \includegraphics[height=0.3\textwidth]{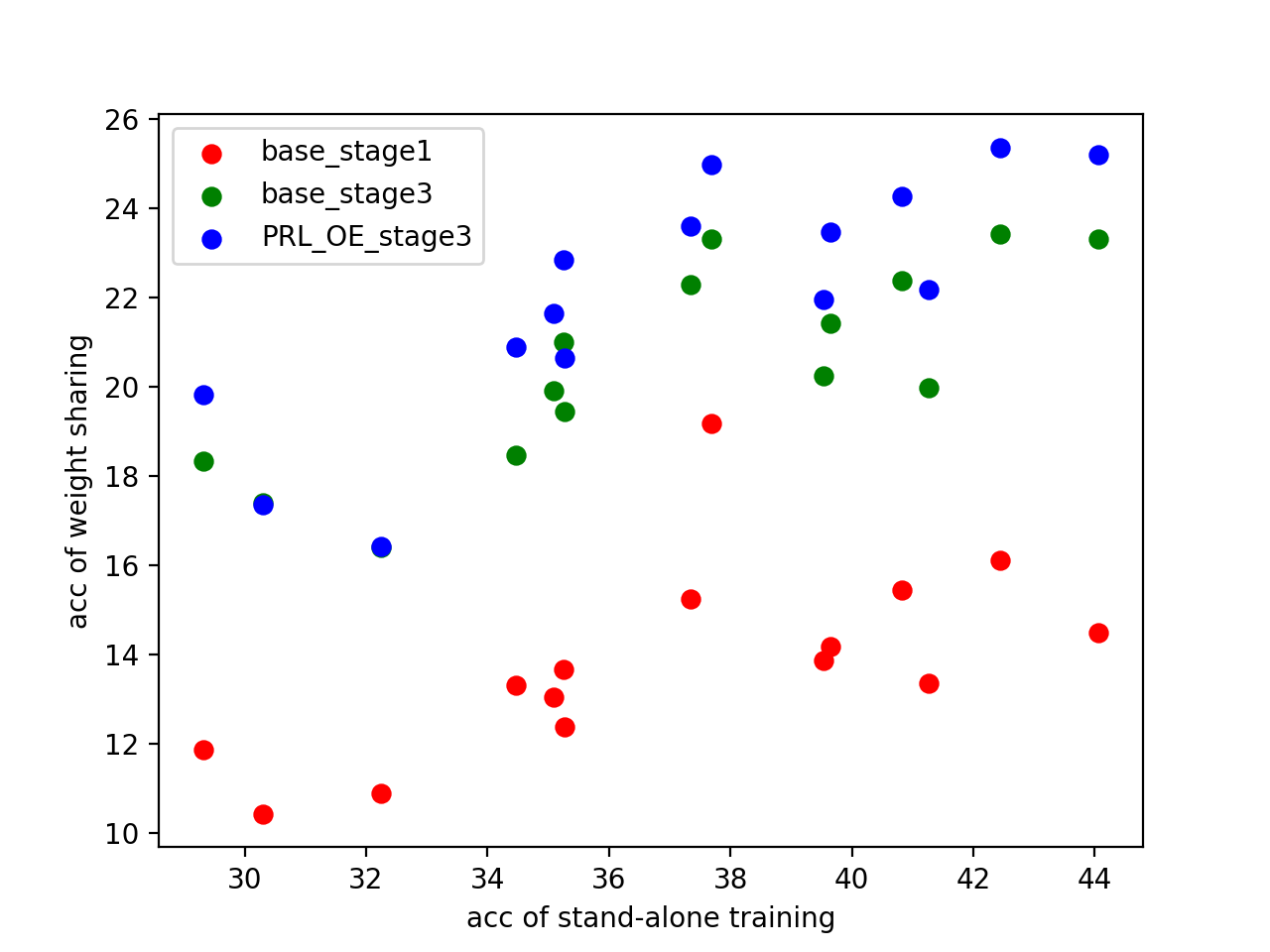}
   \end{center}
      \caption{The accuracy correlation of subet-networks in different training settings.}
   \label{fig:correlation}
\end{figure}

We train a few sub-networks independently and get their accuracies on validation set.
Then the correlation of accuracies obtained with supernet and stand-alone training are displayed in Figure~\ref{fig:correlation},
which visualizes the improvement of accuraies and ranking correlation of sub-networks in the supernet.

\section{Conclusion}
In this paper, we improve the ranking correlation of supernet with capacity enlarging and progressive training pipeline.
Firstly, We utilize simple yet effective channel proxies and activation conversion for candidates enhancing.
Secondly, we gradually add the counterpart of weights for different searchable options and elaborately finetune the supernet, which contributes to better convergence and ranking consistency.
Finally, the experiments demonstrate that both candidates enhancement and progressive training improve the ranking correlation of sub-networks in weights inheriting and stand-alone training.

{\small
\bibliographystyle{ieee_fullname}
\bibliography{egbib}
}

\end{document}